\pdfoutput=1

\documentclass[11pt]{article}

\usepackage[final]{acl}
\usepackage{times}
\usepackage{latexsym}
\usepackage{color}
\usepackage{algorithm,algpseudocode}
\usepackage{amsmath}
\usepackage{amsfonts}
\usepackage{xcolor,colortbl}
\usepackage{amssymb}
\usepackage{algpseudocode}
\usepackage{multirow}
\usepackage{tcolorbox}
\usepackage{ulem}
\usepackage{caption}
\usepackage{subcaption}

\definecolor{lightblue}{RGB}{225, 245, 254}
\definecolor{lightyellow}{RGB}{255, 253, 231}
\definecolor{lightgray}{RGB}{236, 239, 241}
\definecolor{lightgray}{rgb}{0.83, 0.83, 0.83}
\definecolor{lightbrown}{RGB}{239, 235, 233}
\definecolor{lightgreen}{RGB}{241, 248, 233}
\definecolor{lavenderpink}{rgb}{0.98, 0.68, 0.82}
\definecolor{lavenderindigo}{rgb}{0.58, 0.34, 0.92}
\definecolor{brilliantlavender}{rgb}{0.96, 0.73, 1.0}
\definecolor{columbiablue}{rgb}{0.61, 0.87, 1.0}

\algnewcommand\algorithmicforeach{\textbf{for}}
\algdef{S}[FOR]{For}[1]{\algorithmicforeach\ #1\ \algorithmicdo}

\usepackage{times}
\usepackage{latexsym}
\usepackage{eucal}
\usepackage{soul}
\usepackage[T1]{fontenc}
\usepackage{booktabs}
\usepackage[utf8]{inputenc}

\usepackage{microtype}
\usepackage{lipsum}
\usepackage{inconsolata}

\usepackage{inconsolata}

\usepackage[utf8]{inputenc}
\usepackage[T1]{fontenc}
\usepackage{hyphenat}
\usepackage{xspace}
\usepackage{amsmath}
\usepackage{amsfonts}
\usepackage{hyperref}
\usepackage{url}
\usepackage{booktabs}
\usepackage{multirow}
\usepackage{makecell}
\usepackage{caption}
\usepackage{minibox}
\usepackage{bbm}
\usepackage{graphicx}
\usepackage{balance}
\usepackage{mathtools}
\usepackage{color}
\usepackage{marvosym}
\usepackage{ifthen}
\usepackage{textcomp}
\usepackage{enumitem}
\usepackage{verbatim}
\usepackage{algorithm}
\usepackage{numprint}
\usepackage{balance}

\usepackage{amsthm}
\theoremstyle{plain}

\newcommand{\chatoDisplayMode}[1]{#1}

\definecolor{MyRed}{rgb}{0.6,0.0,0.0} 
\definecolor{MyBlack}{rgb}{0.1,0.1,0.1} 
\newcommand{\inred}[1]{{\color{MyRed}\sf\textbf{\textsc{#1}}}}
\newcommand{\frameit}[2]{
  \begin{center}
  {\color{MyRed}
  \framebox[.9\columnwidth][l]{
    \begin{minipage}{.85\columnwidth}
    \inred{#1}: {\sf\color{MyBlack}#2}
    \end{minipage}
  }\\
  }
  \end{center}
}

\newcommand{\note}[2][]{\chatoDisplayMode{\def\@tmpsig{#1}\frameit{{\Pointinghand} Note}{#2\ifx \@tmpsig \@empty \else \mbox{ --\em #1}\fi}}}
\newcommand{\todo}[2][]{\chatoDisplayMode{\def\@tmpsig{#1}\frameit{{\Writinghand} To-do}{#2\ifx \@tmpsig \@empty \else \mbox{ --\em #1}\fi}}}

\newcommand{\textcite}[1]{\citeauthor{#1} \shortcite{#1}}

\newcommand{\hide}[1]{}

\makeatletter
\newcommand{\iffont}[2]{\ifthenelse{\equal{\f@family}{#1}}{#2}{}}
\makeatother

  \usepackage{mathptmx}

  \DeclareSymbolFont{greek}{OML}{cmm}{m}{n}
  \DeclareMathSymbol{\alpha}{\mathalpha}{greek}{"0B}
  \DeclareMathSymbol{\beta}{\mathalpha}{greek}{"0C}
  \DeclareMathSymbol{\gamma}{\mathalpha}{greek}{"0D}
  \DeclareMathSymbol{\delta}{\mathalpha}{greek}{"0E}
  \DeclareMathSymbol{\epsilon}{\mathalpha}{greek}{"0F}
  \DeclareMathSymbol{\zeta}{\mathalpha}{greek}{"10}
  \DeclareMathSymbol{\eta}{\mathalpha}{greek}{"11}
  \DeclareMathSymbol{\theta}{\mathalpha}{greek}{"12}
  \DeclareMathSymbol{\iota}{\mathalpha}{greek}{"13}
  \DeclareMathSymbol{\kappa}{\mathalpha}{greek}{"14}
  \DeclareMathSymbol{\lambda}{\mathalpha}{greek}{"15}
  \DeclareMathSymbol{\mu}{\mathalpha}{greek}{"16}
  \DeclareMathSymbol{\nu}{\mathalpha}{greek}{"17}
  \DeclareMathSymbol{\xi}{\mathalpha}{greek}{"18}
  \DeclareMathSymbol{\pi}{\mathalpha}{greek}{"19}
  \DeclareMathSymbol{\rho}{\mathalpha}{greek}{"1A}
  \DeclareMathSymbol{\sigma}{\mathalpha}{greek}{"1B}
  \DeclareMathSymbol{\tau}{\mathalpha}{greek}{"1C}
  \DeclareMathSymbol{\upsilon}{\mathalpha}{greek}{"1D}
  \DeclareMathSymbol{\phi}{\mathalpha}{greek}{"1E}
  \DeclareMathSymbol{\chi}{\mathalpha}{greek}{"1F}
  \DeclareMathSymbol{\psi}{\mathalpha}{greek}{"20}
  \DeclareMathSymbol{\omega}{\mathalpha}{greek}{"21}
  \DeclareMathSymbol{\varepsilon}{\mathalpha}{greek}{"22}
  \DeclareMathSymbol{\vartheta}{\mathalpha}{greek}{"23}
  \DeclareMathSymbol{\varpi}{\mathalpha}{greek}{"24}
  \DeclareMathSymbol{\varrho}{\mathalpha}{greek}{"25}
  \DeclareMathSymbol{\varsigma}{\mathalpha}{greek}{"26}
  \DeclareMathSymbol{\varphi}{\mathalpha}{greek}{"27}
  \DeclareSymbolFont{otone}{OT1}{cmr}{m}{n}
  \DeclareMathSymbol{\Gamma}{\mathalpha}{otone}{0}
  \DeclareMathSymbol{\Delta}{\mathalpha}{otone}{1}
  \DeclareMathSymbol{\Theta}{\mathalpha}{otone}{2}
  \DeclareMathSymbol{\Lambda}{\mathalpha}{otone}{3}
  \DeclareMathSymbol{\Xi}{\mathalpha}{otone}{4}
  \DeclareMathSymbol{\Pi}{\mathalpha}{otone}{5}
  \DeclareMathSymbol{\Sigma}{\mathalpha}{otone}{6}
  \DeclareMathSymbol{\Upsilon}{\mathalpha}{otone}{7}
  \DeclareMathSymbol{\Phi}{\mathalpha}{otone}{8}
  \DeclareMathSymbol{\Psi}{\mathalpha}{otone}{9}
  \DeclareMathSymbol{\Omega}{\mathalpha}{otone}{10}
  \DeclareSymbolFont{syms}{OML}{cmm}{m}{it}
  \DeclareMathSymbol{\partial}{\mathord}{syms}{"40}
  \DeclareMathAlphabet{\mathbold}{OML}{cmm}{b}{it}
  \DeclareSymbolFont{largesymbols}{OMX}{cmex}{m}{n}

\usepackage{amsmath}
\usepackage{amssymb}

\usepackage{epigraph}
\newcommand{\ourmodel}{\textsc{Frodo}\xspace}
\definecolor{MyBlue}{rgb}{0.25,0.5,0.75}

\definecolor{blush}{rgb}{0.87, 0.36, 0.51}

\colorlet{NextBlue}{MyBlue!20}
\colorlet{NextBlue}{MyBlue!20}
\colorlet{SecondBlue}{MyBlue!40}
\colorlet{Nextblush}{blush!30}

\setul{0.5ex}{0.3ex}
\definecolor{Green}{rgb}{0,1,0}
\setulcolor{Green}

\setul{0.5ex}{0.3ex}
\definecolor{Red}{rgb}{1,0.0,0.0}
\setulcolor{Red}
\usepackage{pifont}
\newcommand{\xmark}{\ding{55}}%

\title{Making Reasoning Matter:\\ Measuring and Improving Faithfulness of Chain-of-Thought Reasoning}

\author{First Author \\
  Affiliation / Address line 1 \\
  Affiliation / Address line 2 \\
  Affiliation / Address line 3 \\
  \texttt{email@domain} \\\And
  Second Author \\
  Affiliation / Address line 1 \\
  Affiliation / Address line 2 \\
  Affiliation / Address line 3 \\
  \texttt{email@domain} \\}

\author{Debjit Paul, 
  \textbf{Robert West},
  \textbf{Antoine Bosselut}, 
  \textbf{Boi Faltings} \\
  EPFL \\
  \texttt{\{debjit.paul, robert.west, antoine.bosselut, boi.faltings}\}@epfl.ch \\
}

\begin{document}
\maketitle
\begin{abstract}



Large language models (LLMs) have been shown to perform better when asked to reason step-by-step before answering a question. However, it is unclear to what degree the model's final answer is faithful to the stated reasoning steps. In this paper, we perform a causal mediation analysis on {twelve} LLMs to examine how intermediate reasoning steps generated by the LLM influence the final outcome and find that LLMs do not reliably use their intermediate reasoning steps when generating an answer. To address this issue, we introduce \ourmodel{}, a framework to tailor small-sized LMs to generate correct reasoning steps and robustly reason over these steps. \ourmodel{} consists of an \textit{inference module} that learns to generate correct reasoning steps using an implicit causal reward function and a \textit{reasoning module} that learns to faithfully reason over these intermediate inferences using a counterfactual and causal preference objective. Our experiments show that \ourmodel{} significantly outperforms four competitive baselines. Furthermore, \ourmodel improves the robustness and generalization ability of the reasoning LM, yielding higher performance on out-of-distribution test sets. Finally, we find that \ourmodel{}'s rationales are more faithful to its final answer predictions than standard supervised fine-tuning.

 
\end{abstract}

\section{Introduction} 




Chain-of-thought (CoT) reasoning techniques have been shown to improve the performance of large language models (LLMs) by generating step-by-step reasoning traces before generating a final answer \citep{wei2022chain}.
Many works suggest that the reasoning process described in CoT explanations may be a possible description of how models make predictions \citep{NEURIPS2022_8bb0d291, yao2023react, sun2023recitationaugmented}. However, despite the remarkable success of CoT in many reasoning tasks, recent works show that LLMs-generated reasoning traces can be incorrect \citep{Zhang2023HowLM} and unfaithful \citep{Turpin2023LanguageMD}. 

\begin{figure}[t]
    \centering   
    \includegraphics[scale=0.9,width=\linewidth]{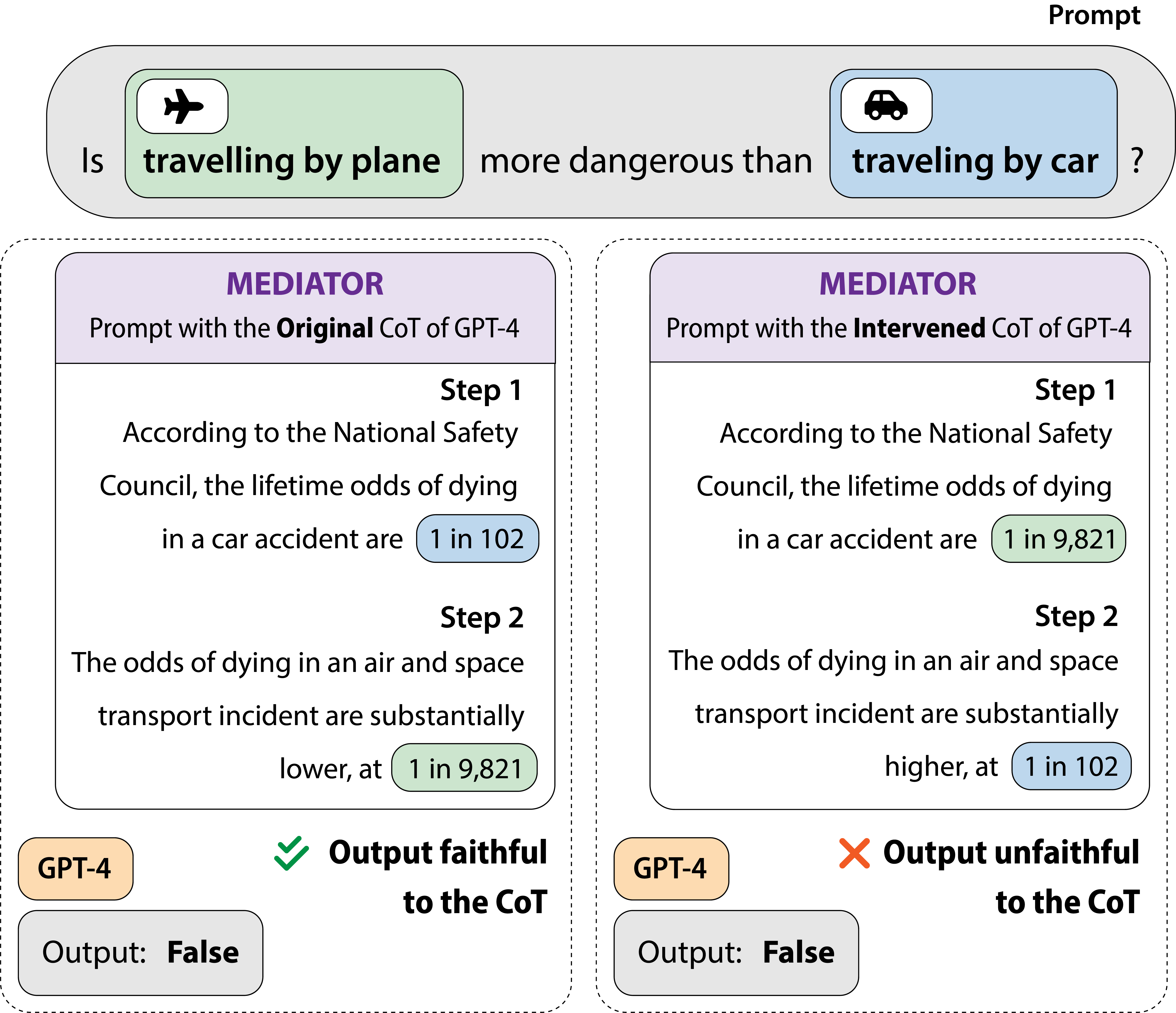}
    \caption{An example of our proposed causal analysis to measure the faithfulness of the final output to the CoT generated by the model. We perturbed CoT rationales and studied the causal impact on the model's behaviour.}
    \label{fig:GPT4_example}
\end{figure} 
Reasoning implicitly involves two steps: identifying the rules and facts (inference chains) necessary to reach a conclusion and then robustly using them to reach said conclusion \citep{levesque1986knowledge}. Our paper studies whether LLMs reliably use inference chains to arrive at a conclusion.\footnote{In our paper, reasoning faithfulness refers to how reliably the model uses its reasoning steps to arrive at a correct answer.} 
In standard CoT, LLMs can generate plausible explanations with the final answer not necessarily guaranteed to follow the reasoning chain or imply a causal relation between the reasoning chain and the model’s outcome \citep{lyu2023faithful}.
Most recent efforts have either focused on the performance of LLMs on various reasoning tasks or their faithfulness in CoT generation, ignoring the sequential relationship between CoT and the final answer \citep{Huang2022TowardsRI, Lanham2023MeasuringFI}.


In this work, we 
address this gap by introducing a methodology for interpreting the relationship between the CoT trace and the final answer based on causal mediation analysis \citep{Pearl2001DirectAI}. Causal mediation analysis is a method of causal inference that studies the change in a response variable following an intervention or treatment. More concretely, we use this method to measure and interpret the contribution of a reasoning chain (mediator) to the final answer (observed output), as shown in Fig.\ref{fig:GPT4_example}. We propose multiple interventions on the model inputs and mediators (reasoning chain) to unveil the causal effect of specific reasoning steps in a model's output. 



We apply this framework and study the causal impact of CoT rationales on the behaviour of twelve different state-of-the-art LLMs on three different complex reasoning tasks (\textit{mathematical, commonsense}, and \textit{causal understanding}). We observe a large variation across tasks and models in how strongly reasoning traces causally affect the model's prediction. In particular, we find that instruction-tuned models (GPT-3.5-Instruct, \citealp{Brown2020LanguageMA}; Mistral-Instruct-7B, \citealp{Jiang2023Mistral7}) have a stronger causal effect on the final answer when conditioned on the reasoning trace than models trained with RLHF (e.g., ChatGPT; Llama-2-7B-Chat, \citealp{Touvron2023Llama2O}). Similar to \citet{Turpin2023LanguageMD}, when we intervene in the reasoning problem, we observe that ChatGPT and GPT-$3.5$-Instruct are inconsistent at generating plausible reasoning chains. Finally, we find GPT-$4$ \citep{Achiam2023GPT4TR} only changes its answer $30$\% of the time when conditioned on perturbed counterfactual reasoning chains. In Figure \ref{fig:GPT4_example}, we see one example where GPT-4 does not faithfully change its final answer when provided with intervened counterfactual CoT. These results indicate two issues: (i) LLMs can generate unfaithful and implausible reasoning chains, and (ii) LLMs are inconsistent when reasoning over their own generated reasoning traces.




To address these issues, we introduce a novel method, \ourmodel{}, comprising two modules. The first module tailors small-sized LMs to generate correct reasoning chains (inference module), while the second module takes the reasoning chains as input and faithfully reasons over them to arrive at the correct answer (reasoning module). To learn to generate correct reasoning chains, we use the DPO algorithm \citep{rafailov2023direct}, which enables the model to prefer correct reasoning chains over counterfactual ones with implicit feedback. Instead of relying on human labeling, we obtain preference data by prompting LLMs to generate correct and counterfactual reasoning chains. Second, we train another small-sized LM to improve the causal effect between the reasoning chain and the final answer using a counterfactual and causal preference ranking objective.

We evaluate \ourmodel on four reasoning tasks (Quarel, StrategyQA, OpenBookQA, QASC) using multiple model backbones of different scales, and demonstrate that \ourmodel achieves an absolute accuracy improvement of $2\%\sim3\%$ over standard supervised fine-tuning or CoT distillation methods. We assess robustness by examining how models alter their answers when intervened with counterfactual reasoning chains. \ourmodel exhibits significant (+$4.5\%$) improvement in robustness. Finally, \ourmodel generalizes better to out-of-distribution test sets, showing a +$2.6\%$ performance improvement over supervised fine-tuning. Our code and data are publicly available\footnote{\url{https://debjitpaul.github.io/reasoningmatter}}.

\section{Reasoning Chain as a Mediator}\label{sec:reasoning_causal}

\paragraph{Problem Formulation.} Reasoning is often a process that involves composing multiple inference steps to reach a conclusion or make a decision. We informally conceptualize each reasoning task as requiring a model $f$: $X$ → $Y$ to map an input $x \in X$ to an output $y \in Y$ by making correct or plausible inference steps $R$. 


\paragraph{Causal Interpretation.} The causal graph is a probabilistic graphical model used to describe how variables interact, expressed by a directed acyclic graph consisting of the sets of nodes ($N$) denoting the variables and a set of directed edges ($E$) indicating the causal relationships between these variables denoting the causality.


\paragraph{Causal Mediation Analysis.} It is a method to measure how an independent variable (or treatment) affects a dependent variable (or outcome) mediated by intermediate variables \citep{Pearl2001DirectAI, robins2003semantics}. 
Causal mediation analysis aims to decompose the total effect of the independent variable ($X$) on the dependent variable ($Y$) into two components: the direct effect and the indirect effect \citep{Pearl2001DirectAI}. In this work, we view the reasoning process as a causal graph, framing the input (reasoning problem) $X$ and the output $Y$ as random variables and the reasoning steps as mediator variable $R$.  We use mediation analysis to interpret the role of reasoning steps as mediators between model inputs and model outputs. 
Let $X_0$ denote the initial reasoning problem, $R_0$ the reasoning chain given $X_0$. Let $Y_{00}$ denote the potential outcome when the treatment and mediator variables are $X_0$ and $R_0$, respectively. Meanwhile, $Y_{01}$ denotes the potential outcome when treatment is set to $X_0$, and $R_1$ is the reasoning chain for the reasoning problem $X_1$. 




\begin{figure}[t]
    \centering   
    \includegraphics[scale=1,height=3cm, width=0.35\paperwidth]{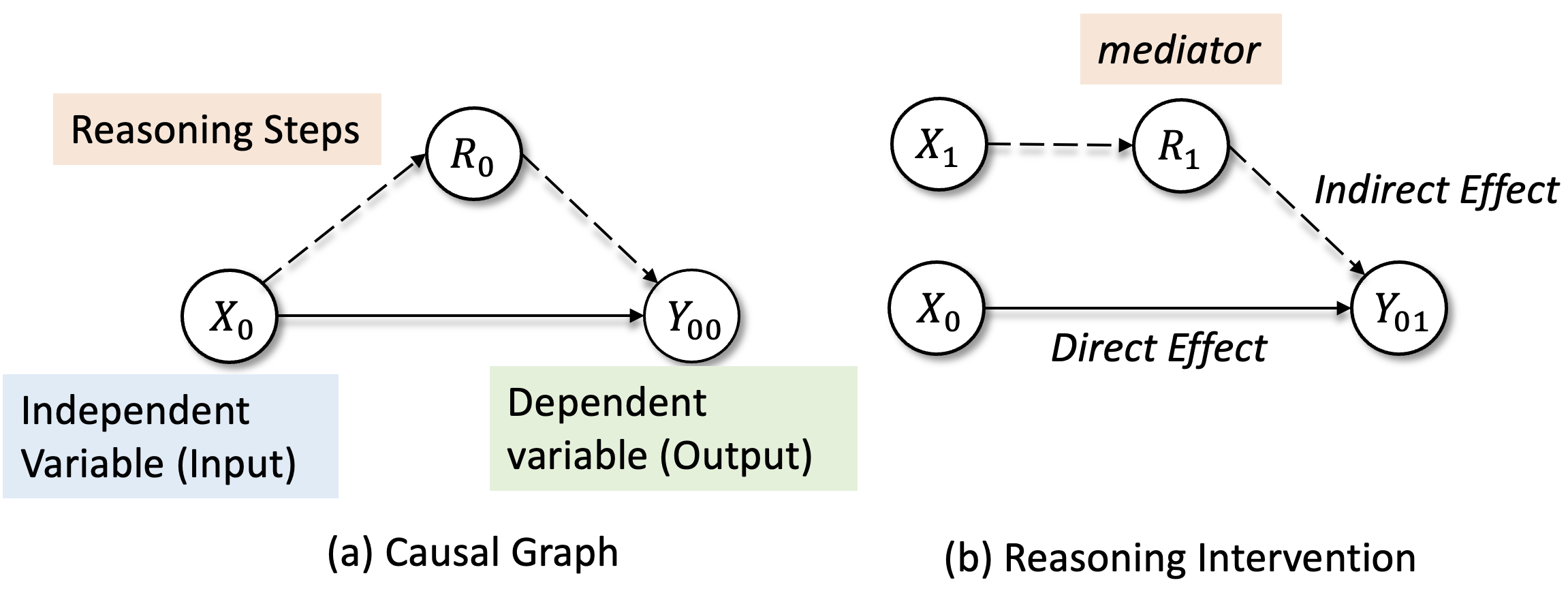}
    \caption{Causal graph for natural language reasoning, modeling $P(Y| do(x))$. $X_0$: original reasoning problem, $X_1$: intervened reasoning problem. $R_0$: Reasoning steps for $X_0$, $R_1$: Reasoning steps for $X_1$. Outputs $Y_{00}$ or $Y_{01}$ are model outputs given $X_0$ and $R_0$ or $X_0$ and $R_1$.}
    \vspace*{-2mm}
    \label{fig:causal_graph}
\end{figure}

\textit{Direct Effect} (DE) \textit{measures how much an intervention $X$ changes an outcome variable $Y$ directly, without passing through a hypothesized mediator $R$}. The direct effect of $X=X_0$ on $Y$ can be defined as $\mathbb{E}[Y_{00} - Y_{10}]$, which can be seen as the correctness comparison between the two potential outcomes given two different treatments, i.e., $X=X_0$ and $X=X_1$. It is computed by applying the intervention $X$ but holding $R$ fixed to its original value ($R_0$). 

\textit{Indirect Effect} (IE) \textit{measures how much an intervention $X$ changes $Y$ indirectly through $R$}. The indirect effect can be defined as IE = $\mathbb{E}[Y_{00} - Y_{01}]$. It is computed by setting $R$ to its value under the intervention $X$ while keeping everything else to its original value.

More concretely, according to \citet{Pearl2001DirectAI}, in our scenario, a high direct effect means the model emphasizes the reasoning problem more than the reasoning steps. In contrast, a high indirect effect means the model emphasizes the reasoning steps more than the problem input.

\paragraph{Reasoning Intervention.} Following \citet{Pearl2001DirectAI}, we conduct counterfactual reasoning to measure the causal effect of a treatment variable on a response variable. We first perform targeted interventions on the input text $X$ and measure their effect on the reasoning outcome $Y$ by keeping $R$ fixed (direct effect). Further, we also perform interventions on the mediator $R$ and measure their effect on $Y$ (indirect effect). We perform the following steps to automatically generate an intervention on $X$ and $R$. \\  
\textbf{Step 1: Intervention Data Generation.}  We use a large language model (GPT-4) to automatically generate an alternative value $X_1$ for the treatment variable.\footnote{See Table. \ref{tab:causal interventions} for details on task-specific interventions.} The input to LLM includes instruction and few-shot examples, taking the format shown in Table \ref{table:counterfactual_question}. LLMs can be sensitive to instructions and few-shot examples; hence, we randomize the prompt by manually creating a set of semantically similar instructions. Then, we randomly sample from the instruction set each time. \\
\textbf{Step 2: Manual Data Curation.} To retain high-quality data for our analysis, we manually filter out generated samples from Step 1 that are invalid or low-quality. Table \ref{tab:example_2} shows an example where given the original input reasoning question $X_0$, the model generated $X_1$, where it replaces ``\textit{Vulcan}'' with ``\textit{Neptune}''. \\ 
\textbf{Step 3: Generate Reasoning Chain.} Finally, to get the indirect effect, we generate the reasoning chain ($R_0$, $R_1$) for each reasoning problem $X_0$ or $X_1$ by providing LLMs with some high-level descriptions about each reasoning task and reasoning prompt -- ``\textit{Let’s think step by step}''(see App. Table. \ref{table:counterfactual_prompt}). 



\setulcolor{Red}
\begin{table}[t!]
\centering
\scalebox{0.7}{
\begin{tabular}{@{}c@{~~}l@{~~}}
\toprule
{\bf Variables } & {\bf Example}\\\midrule
 
$X_0$ & \textit{Is Poseidon similar to the god \ul{Vulcan?}} \\ 
$R_0$ & \textit{Poseidon is a god from Greek mythology, known} \\ 
& \textit{as the \ul{god of the sea}, earthquakes, and horses.} \\
& \textit{Vulcan is a god from Roman mythology, known} \\ 
& \textit{as the \ul{god of fire}, metalworking, and the forge.} \\
& \textit{Although both are gods, they represent different} \\
& \textit{elements and aspects, and come from} \\
& \textit{different mythologies.} \\
\hline
$X_1$ & \textit{Is Poseidon similar to the god \ul{Neptune}?} \\ 
$R_1$ & \textit{Poseidon is a god from Greek mythology, known} \\
& \textit{as the \ul{god of the sea}, storms, and earthquakes.} \\ 
& \textit{Neptune is a god from Roman mythology, who is}  \\ 
& \textit{also known as the \ul{god of the sea}. Both Poseidon and}\\
& \textit{Neptune share similar roles and attributes in} \\
& \textit{their respective mythologies.} \\  
\bottomrule
\end{tabular}}
\caption{An example from StrategyQA dataset, where $X_1$ = intervened reasoning problem; $R_0$ and $R_1$ = reasoning steps (generated by GPT-4).}

\label{tab:example_2}
\end{table}


Our study suggests that vanilla LMs (<20B) (in a zero-shot setting) are systematically unfaithful and consistently fail to reason over the mediator (see Table \ref{tab:open_models_causalresuts}). In general, our experiments show a large variation in the causal effects of COT in the final answer depending on the tasks. Models that are instruction tuned or trained on the chain of thought during the pre-training phase have a better indirect effect across different reasoning tasks, suggesting that fine-tuning on CoT can make the model more faithful (see Table  \ref{tab:closed_models_causal_resuts}). Interestingly, similar to \citet{Turpin2023LanguageMD}, we observe an inverse scaling for certain tasks. In our case, the indirect effect is worsening with increasingly capable models, indicating that smaller models might be more effective in faithful reasoning. 

\section{FRODO} 

In this section, we introduce \ourmodel{}, a framework that tailors small-sized LMs (<10B parameters) to be strong rationalizers and perform reasoning faithfully over the rationales. \ourmodel{} aims to improve the synergy between the reasoning chain and the final answer. We first describe how we obtain silver reasoning chains from LLMs ($\S$\ref{sec:inference_module}). Then, we introduce our inference module that trains a model to generate rationales ($\S$\ref{sec:inference_module}) followed by the reasoner module and its training objectives ($\S$\ref{sec:reasoning_module}).

\subsection{Inference Module}\label{sec:inference_module}
In this work, we assume no gold rationales to train our model. Hence, similar to recent works \citep{liu-etal-2022-rainier, liu-etal-2023-crystal, wang-etal-2023-scott, ramnath2024tailoring}, we automatically obtain the silver rationale from LLM (GPT-3) using in-context learning. A common approach is fine-tuning a smaller text-to-text model on the silver rationales generated by LLMs with a standard language modeling loss. Recent studies have shown that fine-tuning models (<5B) on reasoning chains may struggle to \textit{align} the reasoning chains with the provided reasoning question during inference \citep{Yang2023ChainofThoughtIN, pmlr-v202-fu23d}. Additionally, learning to generate a reasoning chain means learning to decompose complex reasoning into smaller reasoning steps implicitly. However, \citet{shridhar-etal-2023-distilling} showed that fine-tuning could lead to learning shortcuts and degrade performance. Recent studies have demonstrated that feedback-based methods can help the model align better with the human goal. Hence, we use Direct Preference Optimization (DPO) \citep{rafailov2023direct} for aligning LMs to learn to generate correct reasoning chains.

\paragraph{Preference Data.} We prompt the LLM to generate correct reasoning chains ($R_w$) and incorrect reasoning chains ($R_l$) for each reasoning problem. In our experiments, we consider two kinds of reasoning chains as incorrect: \textit{counterfactual chains} (alternative chains that can lead to different outcomes) and \textit{irrelevant chains}. We assume that models that can understand and learn to prefer correct reasoning chains over counterfactual chains will become more robust and enhance generalization. Hence, we manually construct correct and incorrect intermediate reasoning steps and demonstrate the model with these annotated examples before a new instance is provided. In this way, we obtain a preference data $D \in \{X, R_w, R_l\}$ that contains reasoning problems ($X$) and pairs of reasoning steps that lead to correct ($Y_w$) or incorrect outcomes ($Y_l$).

\paragraph{Training.} Given a reasoning problem \{$x \in X$\} and instruction prompt {$p \in$ \{correct or counterfactual}\}, our goal is to train models that could generate reasoning steps ($r_w$ or $r_l$). We propose to adopt Direct Preference Optimization (DPO) \citep{rafailov2023direct}, an effective algorithm for aligning language models with implicit rewards. DPO assumes that we only have access to some pairwise preference data $x \rightarrow$ \{$r_w$ $>$ $r_l$\} for each problem $x \in X$. Hence, while training a model ($\pi_\theta$) to generate correct reasoning steps, we consider counterfactual and irrelevant reasoning steps as less preferred. 
Training a DPO model includes two phases: (i) supervised fine-tuning (SFT) and (ii) Preference Learning (PL) phase.  

\paragraph{SFT.} We begin by fine-tuning a pre-trained LM with a maximum log-likelihood objective to obtain $\pi_{sft}$. 

\paragraph{PL Phase.} Contrary to traditional RL approaches, which initially train a reward model and subsequently derive a policy from it, DPO enables extracting policy through implicit reward learning. DPO adopts a binary classification loss: 
\begin{equation}
        \mathcal{L}_{\text{DPO}} = -\mathbb{E}_{\{x,r_w>r_l\}}\text{log}\sigma(f_\theta (r_w,x)  - f_\theta(r_l,x))
\end{equation}
where $f_\theta$ is the implicit reward model. Intuitively, the gradient of the loss function $\mathcal{L}_{\text{DPO}}$ increases the
likelihood of the preferred completions $r_w$ and decreases the likelihood of counterfactual reasoning chains $r_l$. See Appendix \ref{label:dpo} for more details.  
During inference, the reasoning module uses the generated reasoning steps by  $\pi_\theta$ model for a given reasoning problem. 
\begin{figure}[t]
    \centering   
    \includegraphics[scale=1,height=6cm, width=\linewidth]{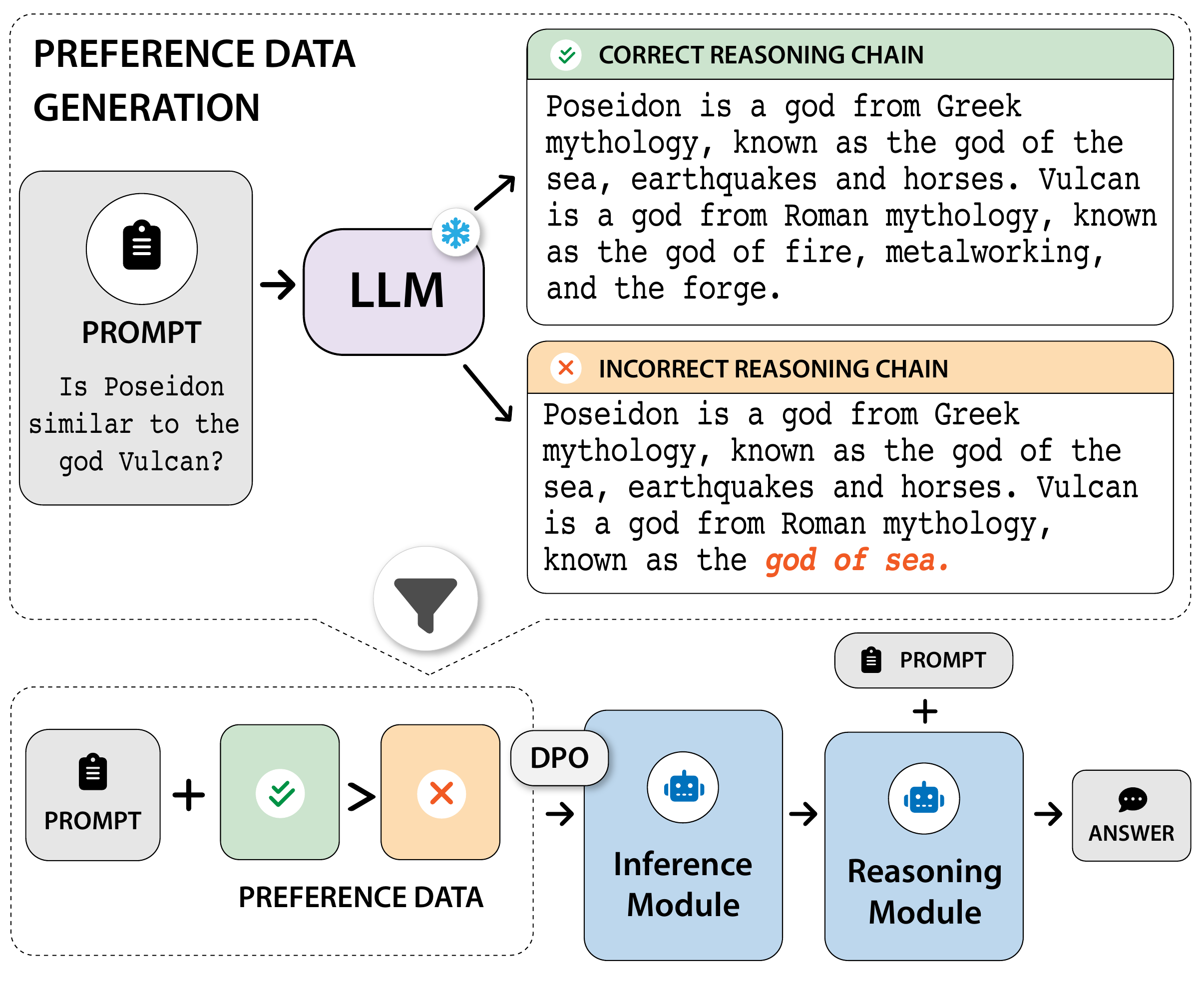}
    \caption{An overview of \ourmodel{}.}
    \label{fig:main_method}
    \vspace*{-5mm}
\end{figure} 

\subsection{Reasoning Module}\label{sec:reasoning_module} Given a reasoning question $x \in X$ and reasoning steps $r_w$ (correct) and $r_l$ (counterfactual)\footnote{Please note that in the reasoner module, we only consider counterfactual reasoning steps as negative samples.}, our goal is to train a model ($\pi_\gamma$) that can generate a correct answer $y_w$. To encourage our reasoner module to reason faithfully over the reasoning steps, we train the model with a linear combination of three losses: an indirect effect loss and a supervised margin rank loss, $\mathcal{L}= 
\lambda_\text{LM}* \mathcal{L}_{\text{LM}} +  
\lambda_\text{counter}*\mathcal{L}_{\text{counter}} +
\lambda_\text{PREF} *\mathcal{L}_{\text{PREF}} $
, which we describe below.

\paragraph{Language Model Loss.} We use the standard training objective to maximize the likelihood of the correct answer using cross-entropy loss, computed as:
\begin{equation}
    \mathcal{L}_{\text{LM}} = -\text{logP}(y_w|x,r_w)
\end{equation}

\paragraph{Counterfactual Loss.} To encourage the model to reason robustly and faithfully towards the reasoning steps, we propose training the model to learn how different reasoning chains (correct or counterfactual) can lead to different outcomes. Hence, inspired by the causal mediation theory \citep{Pearl2001DirectAI}, we use the following loss: 
\begin{equation}
     \mathcal{L}_{\text{counter}} = - {\text{logP}(y_l|x,r_l)}
\end{equation}

Similar to \citep{wang-etal-2023-scott, Roese1997CounterfactualT}, we posit that adding a counterfactual objective can help the model to avoid learning reasoning shortcut between a question and the gold answer since now the model is tasked to answer differently for the same question.

\paragraph{Margin-Ranking Loss.} It has been shown \citep{khosla2020supervised} that contrastive loss and ranking loss help to improve model robustness and generalization against input variation. Hence, we propose to use the margin ranking loss that aims to maximize the margin between positive examples (i.e., statements containing questions, correct reasoning steps and correct answers) and negative examples (i.e., statements containing questions, counterfactual reasoning steps and correct answers). 


\begin{equation}
     \mathcal{L}_{\text{PREF}}= max(0, t*\text{IE}+{m})
\end{equation}

where $t$ is the label (indicating which sample in the pair is better)=$1$, $m$ is the margin=$1.0$, and the indirect effect IE = $h(x,r_w,y_w) - h(x,r_l,y_w)$ where $h$ is the logits. 


\begin{table*}[t!]
\centering
\footnotesize
\scalebox{1.0}{
\begin{tabular}{lcccccccccccc}
\toprule 
  \textbf{Models}
 & \multicolumn{3}{c}{{\textbf{StrategyQA}}} 
 & \multicolumn{3}{c}{{\textbf{GSM8k}}}
 & \multicolumn{3}{c}{{\textbf{Causal Understanding}}}
   \\
 & \multicolumn{1}{c}{CoT (\%)} & \multicolumn{1}{c}{NIE} & \multicolumn{1}{c}{NDE} 
 & \multicolumn{1}{c}{CoT (\%)} & \multicolumn{1}{c}{NIE} & \multicolumn{1}{c}{NDE} 
 & \multicolumn{1}{c}{CoT (\%)} & \multicolumn{1}{c}{NIE} & \multicolumn{1}{c}{NDE} &  \\ 

\midrule
\textbf{ChatGPT}  & 69.2  & 15.3 & 9.1 & 70.1 & 56.3 & 1.01 & 58.8 & 21.1 & 27.4 \\ 
\textbf{GPT-4}    & 93.5 & 40.0 & 22.2 & 81.1 & 21.01 & 30.01 & 72.5 & 29.1  & 48 \\ 

\bottomrule
\end{tabular}}
\caption{\textbf{Causal Effects} of generated CoT and reasoning problems on the outputs, with both Natural Indirect Effect (NIE) and Natural Direct Effect (NDE). COT (\%) represents the accuracy of the models. }\label{tab:closed_models_causal_resuts}

\end{table*}
\section{Experiments}\label{sec:experiments}


\paragraph{Datasets.} We conduct the causal mediation analysis on three datasets: \textsc{StrategyQA} \citep{geva-etal-2021-aristotle}, GSM8K \citep{cobbe2021gsm8k}, and Causal Understanding \citep{suzgun-etal-2023-challenging}. 
We evaluate \ourmodel{} on four datasets: \textsc{StrategyQA}, \textsc{Quarel}\citep{Tafjord2018QuaRelAD}, \textsc{OpenBookQA} \citep{mihaylov-etal-2018-suit}, and \textsc{QASC} \citep{khot2020qasc}.  We report more details about each dataset in App. \ref{sec:appendix_data_details}. For all the datasets, we do not use human-written rationales. We used rationales generated by prior work \citep{ramnath2024tailoring} using GPT-3 (\textsc{Text-Davinci-003}) as silver rationales for supervision. For counterfactual rationales, we use chain-of-thought prompts on these datasets (Table. \ref{table:counterfactual_prompt}) and sample 2 rationales for each training instance with a temperature of 0.5. 

\paragraph{Evaluation Metrics.} To evaluate the causal effects, we report the average indirect and direct effects of the LLMs. We use the following formula to calculate the scores: $IE$ = Avg[$\text{Acc}(Y_{00}) - \text{Acc} (Y_{01})$], and $DE$ = Avg[$\text{Acc}(Y_{00}) - \text{Acc}(Y_{10})]$ where $X_0$ and $R_0$ original reasoning problem and reasoning chains. We measure two kinds of causal effects: \textbf{natural} and \textbf{controlled} for different types of LLMs. \textit{Natural Indirect Effect (NIE)}: for models that have emergent capabilities (>$100$B parameters) of generating plausible reasoning chains, we measure the causal effect of $X$ on $Y$ that uses $R$ generated by the same model. \textit{Controlled Indirect Effect (CIE)}: for models with <$20$B parameters, we evaluate the causal effect by providing reasoning chains generated by GPT-4. Further, to measure the \textit{robustness} of models, we use controlled indirect effect. To evaluate the \textit{faithfulness} of the rationales generated by the small-sized models, we use LAS \citep{hase-etal-2020-leakage} to measure how well the rationales help a simulator to predict a student’s predictions a', namely \text{Acc}(qr --> a') - \text{Acc}(q --> a'). Similar to \citet{wang-etal-2023-scott}, we implement each simulator with a fine-tuned T5-large model respectively. 

\paragraph{Implementation Details.}  We use GPT-4 to generate intervened reasoning problems $X_1$ and reasoning chains ($R_0$ or $R_1$) to perform the causal mediation analysis. We report the prompts used in Table.\ref{table:counterfactual_prompt}, \ref{table:counterfactual_question} and hyperparameters in App. \ref{sec:appendix_data_details}. 

\paragraph{Baselines.} We perform the causal analysis on a series of language models that are diverse in terms of scale, training, and data: LLaMa-2 \citep{Touvron2023Llama2O}, Mistral \citep{Jiang2023Mistral7}, Chat-GPT \citep{NEURIPS2020_gpt3}, GPT-4 \citep{openai2023gpt4}, Flan-T5 \citep{chung2022scaling}, Flan-Alpaca \citep{chung2022scaling}, Stable-Vicuna \citep{vicuna2023}. We compare \ourmodel{} with four strong baselines: (1) SFT + CoT: Finetuning a T5-large or T5-3B or LLaMa-2-7B with LoRA or Mistral-7B with LoRA on silver rationales, then train another model with LM objective to perform the reasoning, (2) Rainier \citep{liu-etal-2022-rainier}, where they used PPO (Proximal Policy Optimization) inference modules, and for the reasoning module, they used SFT (simple finetuning). (3) Crystal \citep{liu-etal-2023-crystal} used PPO to train both inference and reasoning modules, (4) Mario \citep{ramnath2024tailoring} used QUARK, a multi-reward reinforcement learning method, and for the reasoning module, they used SFT and (5) SCOTT\citep{wang-etal-2023-scott} used simple-finetuning with contrastive decoding. More details about all the baselines are reported in App. \ref{sec:appendix_baselines}.

\begin{table*}[t!]
\centering
\footnotesize
\scalebox{0.9}{
\begin{tabular}{llccccccccc}
\toprule 
  {} & \textbf{Models}
 & \multicolumn{3}{c}{{\textbf{StrategyQA}}} 
 & \multicolumn{3}{c}{{\textbf{GSM8k}}}
 & \multicolumn{3}{c}{{\textbf{Causal}}}
   \\
 & & {CIE} & {CDE} & \textbf{p-value}
 & {CIE} & {CDE} & \textbf{p-value}
 & {CIE} & {CDE} & \textbf{p-value} \\ 
\midrule
AR & \textbf{LLaMA-2-7B}  &  24.5 & 25 & <0.001 & 27.5 & 8.5 & <0.001 & 2.3 & 1.1 & <0.005\\ 
 & \textbf{Mistral-7B} & 21.2 & 17.9 & <0.001 & 25.1 & 3.8 & <0.001 & 2.3 & 0.6 & <0.009 \\
 & \textbf{LLaMA-3-70B} & 26.6 & 30.6 &  <0.001 & 57.2 & 5.2 & <0.005 & 8.0 &  5.1& <0.002\\
\midrule 
In-context & \textbf{LLaMA-2-7B}  & 24.9 & 10 & <0.005 & 45.6 & 0.9  & <0.005 &  5.6  & 5.6 & <0.009 \\  
\midrule 
MoE & \textbf{Mixtral-8-7B} & 21 & 11 & <0.001 & 47.4 & 2.9 & <0.003 & 5.1 & 4.6 & <0.001\\ 
\midrule 
RLHF & \textbf{LLaMA-2-7B-Chat} & 8.4 & 30.5 & <0.010 & 1.4 & 36.7 & <0.010 & -2.3 & 8.6 & <0.016\\ 
& \textbf{Stable Vicuna-13B} & 3.5 & 2.5 & <0.001 & 45.1 & 2.4 & <0.010 & 0.6 & 0.1 & <0.010\\
& \textbf{ChatGPT} & 2.6 & 13.6  & <0.016  & 57.8 & 16.6 & <0.010& 4.6 & 10.8 & <0.001\\
\midrule 
Instruct Tuned & \textbf{Mistral-Instruct-7B} & 31.6 & 31.9 & <0.001 & 35.5 & 4.7 & <0.001 & 7.4 & 8 & <0.005 \\
\midrule
RLHF + Instruct Tuned & \textbf{GPT-3.5-Instruct} & 26.1 & 27.3 & <0.005 & \textbf{62.6} & 14.7 & <0.005 & \textbf{8.5} & 10.7 & <0.005\\ 
\midrule 
Instruct-Tuned +  & \textbf{Flan-T5-11B} & \textbf{36.9} & 35.7 & <0.001 & 31.23 & 12.2 & <0.001 & 7.4 & 13.1 & <0.001\\ 
CoT Tuned & \textbf{Flan-Alpaca-11B} & 31.2 & 47.9 & <0.001 & 25 & 7.9 &  <0.001& 3.4 & 9.2 & <0.001\\ 

\bottomrule
\end{tabular}}
\caption{\textbf{Causal Effects} of CoT. The reported results are zero-shot performance. CIE: Controlled Indirect Effect, CDE: Controlled Direct Effect. 
The p-value represents the significance of the results}\label{tab:open_models_causalresuts}

\end{table*} 

\section{Results}\label{sec:result}



In Table. \ref{tab:closed_models_causal_resuts}, \ref{tab:open_models_causalresuts}, we report the results of the causal mediation analysis for twelve models. In section \S\ref{sec:experiments}, we provide the details about the implementation, evaluation metrics and datasets. 

\paragraph{Natural Direct and Indirect effects.} We first evaluate the indirect and direct effects of the reasoning chain and reasoning problems on the final outputs. For models (>100B) with the emergent ability to generate plausible reasoning chains, we report \textit{natural} indirect effects and direct effects (see \S\ref{sec:reasoning_causal}). Table \ref{tab:closed_models_causal_resuts} shows the zero-shot performance of the ChatGPT and GPT-4 models. We observe that for StrategyQA and Causal Understanding tasks, GPT-4 has a higher natural indirect effect than ChatGPT, suggesting that it is able to better reason over the reasoning steps for these tasks. However, for mathematical reasoning (GSM8K), ChatGPT has a better indirect effect. Qualitatively, we find that for mathematical reasoning, when we provide intervened reasoning steps, GPT-4 considers them incorrect and continues to generate correct reasoning steps. This results in a lower indirect effect score. Moreover, GPT-4 exhibits a more pronounced direct effect than ChatGPT, suggesting that its outputs are more causally sensitive to reasoning problems. In general, our experiments show a large variation in the causal effects of COT in the final answer depending on the tasks. 

\paragraph{Controlled Direct and Indirect effects.} Table \ref{tab:open_models_causalresuts} shows the results of causal mediation analysis for 12 different LMs. In these experiments, we examined the causal behaviour using reasoning chains generated by GPT-4 (controlled setting). Our study suggests that vanilla LMs (<20B) (in a zero-shot setting) are systematically unfaithful and consistently fail to reason over the mediator. Increasing the model size (7B to 70B) improves the indirect effect (makes them more faithful), indicating the importance of model size. We find that in-context learning and instruction-tuning improve the indirect effect over models trained only with language modelling objectives (e.g., LLaMA and Mistral), indicating that these methods help the model align better with the reasoning chains. We observe that models trained with RLHF objective (ChatGPT, Llama-2-7B-Chat) have a more direct effect than an indirect effect, suggesting that training on human feedback might have disincentive faithful reasoning \citep{sharma-2023-discovering}. Models that are instruction-tuned or trained on the chain of thought (e.g., Flan-T5) during the pre-training phase have a better indirect effect across different reasoning tasks, suggesting that fine-tuning on CoT can make the model more faithful. Similar to \citet{Turpin2023LanguageMD}, we observe inverse scaling for certain tasks. In our case, the indirect effect worsens with increasingly capable models, indicating that sheer scale might not guarantee faithful reasoning. Interestingly, we also observe that none of the models has high indirect or direct effects on the causal understanding task. One intuitive reason is that the causal understanding task is challenging, and the model's (<10B) performance is nearly random; hence, the effects are not strong. Overall, we observe that LLMs are inconsistent in faithfully performing reasoning over the CoT.

\begin{table}[t!]
\centering
\footnotesize
\scalebox{0.95}{
\begin{tabular}{lllll}
\toprule 
  \textbf{Models}
 & \textbf{StrategyQA}
 & \textbf{QuaRel}
 & \textbf{OBQA} 
 & \textbf{QASC}  \\
 \midrule
\textbf{GPT-$3.5^{\diamond}$}  &  69.7  & 83.4 & 84.5 &  80.3 \\ 
\midrule
\textbf{SFT}  &   \cellcolor{teal!25}57.6 &  \cellcolor{teal!25}74.6   & \cellcolor{teal!25}65.0 &   \cellcolor{teal!25}58.6 \\ 
\textbf{SFT + CoT}  &  \cellcolor{teal!25}63.6 & \cellcolor{teal!25}77.7 & \cellcolor{teal!25}65.5 & \cellcolor{teal!25}59.4 \\ 
\textbf{Rainier}  & \cellcolor{teal!25} --   &  \cellcolor{teal!25}-- &  \cellcolor{teal!25}69.7 & \cellcolor{teal!25}54.9 \\ 
\textbf{Crystal}  &   \cellcolor{teal!25}-- &  \cellcolor{teal!25}-- &  \cellcolor{teal!25}64.2 & \cellcolor{teal!25}56.8 \\ 
\textbf{\textsc{Mario}}  &   \cellcolor{teal!25}65.1  &  \cellcolor{teal!25}79.9 & \cellcolor{teal!25}66.1  & \cellcolor{teal!25}60.1 \\ 
\textbf{\ourmodel{}}  &  \cellcolor{teal!25}\textbf{68.4}$^*$  &  \cellcolor{teal!25}\textbf{83.4}$^*$ & \cellcolor{teal!25}70.2$^+$  & \cellcolor{teal!25}\textbf{64.2}$^*$\\ 
\textbf{-DPO} &  \cellcolor{teal!25}{66.2}  &  \cellcolor{teal!25}{82.2} &  \cellcolor{teal!25}68.1 & \cellcolor{teal!25}{62.4} \\
\textbf{-CL} &  \cellcolor{teal!25}{65.2}  &  \cellcolor{teal!25}{82.1} & \cellcolor{teal!25}66.4  & \cellcolor{teal!25}{60.1} \\ 
\textbf{-MRL} &  \cellcolor{teal!25}{65.5}  &  \cellcolor{teal!25}{81.3} &\cellcolor{teal!25}66.2  & \cellcolor{teal!25}{62.1}\\ 
\midrule
\textbf{SFT}  &  \cellcolor{pink!25}63.1 & \cellcolor{pink!25}81.29 &	\cellcolor{pink!25}72.0 &	\cellcolor{pink!25}67.8 \\ 
\textbf{SFT + CoT}  & \cellcolor{pink!25}65.1	& \cellcolor{pink!25}84.2	& \cellcolor{pink!25}73.3	& \cellcolor{pink!25}72.0 \\ 
\textbf{SCOTT} & \cellcolor{pink!25}{61.5}&\cellcolor{pink!25} --& \cellcolor{pink!25}-- & \cellcolor{pink!25}{65.0} \\
\textbf{Crystal}  &  \cellcolor{pink!25}-- &  \cellcolor{pink!25}-- &  \cellcolor{pink!25}78.3 & \cellcolor{pink!25}74.3 \\ 

\textbf{\ourmodel{}}  & \cellcolor{pink!25}\textbf{82.1}$^*$ & \cellcolor{pink!25}\textbf{93.5}$^*$ &  \cellcolor{pink!25}\textbf{80.1}$^*$  &  \cellcolor{pink!25}\textbf{75.9}$^*$ \\
\midrule
\textbf{LlaMa-2-7B}  & \cellcolor{yellow!25}67.2  &\cellcolor{yellow!25}56.8& \cellcolor{yellow!25}47.5 &\cellcolor{yellow!25}49.6 \\ 
\textbf{SFT + CoT}  & \cellcolor{yellow!25}79.4 & \cellcolor{yellow!25}68.4 &\cellcolor{yellow!25}62.8 & \cellcolor{yellow!25}54.6\\
\textbf{\ourmodel{}} & \cellcolor{yellow!25}81.5$^+$ &  \cellcolor{yellow!25}73.5$^+$ \cellcolor{yellow!25} & \cellcolor{yellow!25}71.4$^+$ & \cellcolor{yellow!25}63.4$^+$\\
\midrule
\textbf{Mistral-7B}  &  \cellcolor{cyan!25}58.2  & \cellcolor{cyan!25}56.8 &\cellcolor{cyan!25}82.1 &\cellcolor{cyan!25}65.2\\ 
\textbf{SFT + CoT}  & \cellcolor{cyan!25}78.2 &  \cellcolor{cyan!25}70.8& \cellcolor{cyan!25}83.5 & \cellcolor{cyan!25}70.1\\
\textbf{\ourmodel{}} & \cellcolor{cyan!25}81.9$^+$ &\cellcolor{cyan!25}78.2$^+$  &\cellcolor{cyan!25}84.9$^+$& \cellcolor{cyan!25}72.3$^+$\\
\bottomrule
\end{tabular}}
\caption{Performance of small-sized LMs (770M-7B) on four different reasoning tasks. The base models are \colorbox{teal!25} {T5-large} (770M),  \colorbox{pink!25}{T5-3B} (3B), \colorbox{yellow!25}{LLaMa-2-7B} and \colorbox{cyan!25}{Mistral-7B} . We report accuracy (\%).{$^\diamond$}: few-shot performance, $^*$: p-value<0.01, $^+$: p-value<0.05 }\label{tab:main_resuts}

\end{table}

\textbf{Comparing \ourmodel{} with Baselines.} We now empirically compare \ourmodel{} with three strong baseline models (see Table \ref{tab:main_resuts}). We consider T5-large (770M) as the inference and reasoning modules. We have the following three observations. First, we present the performance of GPT-$3.5$ on these tasks. We observe the performance on StrategyQA is much lower than on other tasks, indicating the rationales generated for this task can be unfaithful. Hence, similar to \citep{ramnath2024tailoring}, for training \ourmodel{}, we use only the instances where the answer predicted by GPT-$3.5$ is correct. Second, for all four datasets, we observe that \ourmodel{} outperforms the strong self-rationalization baselines. \ourmodel{}, on average, improves the performance by +$4.1$ and +$3$ accuracy points compared to the SFT + CoT and \textsc{Mario} (the strongest baseline), respectively, across all four tasks. Since SFT + CoT and \textsc{Mario} use the same knowledge from GPT-3.5, our results suggest that both our inference and reasoning modules bring substantial performance gains to the model. Third, it is worth noting that increasing (770M to 3B) the model size does not hamper the performance of \ourmodel{}. Fourth, we also report the performance of the LLaMa-2-7B and Mistal-7B models. We show that FRODO further improves the performance of model size 7B.

\begin{table}[ht!]
\centering
\footnotesize
\scalebox{1.0}{
\begin{tabular}{lcc}
\toprule 
  \textbf{Models}
 & {\textbf{StrategyQA}}
 & {\textbf{QuaRel}} \\
\midrule
\textbf{SFT}  & 19.4 & 19.4  \\
\textbf{SFT + CoT}  & 32.2 & 29.2  \\
\textbf{\ourmodel{}}  & \textbf{39.9}  &  \textbf{31.2} \\ 
\textbf{-CL} & 34.6 & 28.7 \\
\textbf{-MRL} & 36.2 & 30.6\\
\bottomrule
\end{tabular}}
\caption{\textbf{Robustness Performance} of LLMs on Reasoning over a Chain. We report CIE scores.}\label{tab:robustness}
\end{table}
\paragraph{Ablation.} To obtain a better insight into the contribution of each component of \ourmodel{}, we perform an ablation study (see Table. \ref{tab:main_resuts}). First, when we do not use the DPO to train our inference module, we see a consistent drop (-$1.9\%$) in performance across the four tasks, indicating the importance of incorporating implicit feedback provided by the DPO in the model's training. Further, we observe a considerable drop in performance when we do not use counterfactual (-$3.1\%$) and margin ranking loss (-$2.8\%$). This result highlights the model's ability to benefit from including counterfactual examples. \\
\vspace*{-5mm}



\section{Analysis} 
\subsection{Quantitative Analysis}
\paragraph{Robustness.} In Table \ref{tab:robustness}, we report the controlled indirect effect that indicates how robustly models are able to change their answers when provided with controlled (generated by GPT-4) counterfactual reasoning chains. For \textsc{StrategyQA}, we observe that \ourmodel{} significantly improves the robustness performance for T5-3B (+7.7 pp.). Further, for the QuaRel task, we observe +2 pp. improvement over the SFT + CoT method. Qualitatively, we find that for the MCQA tasks, the gold rationales often contain the answer; hence, the SFT + CoT learns to copy those as answers. Further, we perform an ablation to understand which component contributes most to the model's robustness. We find that counterfactual loss brings the most gain in robustness. 
\paragraph{Generalization.} The idea is to test our model's capability to determine if it can improve out-of-distribution (OOD) generalization. Table \ref{tab:generalization} shows the OOD performance, where we compare our method with SFT+CoT. We trained the models on the OBQA and QASC datasets and evaluated them on the StrategyQA task. We conclude that \ourmodel{} significantly helps improve the model's generalizability to a dataset unseen during fine-tuning. \\ 

\vspace*{-5mm}
\subsection{Qualitative Analysis}
\paragraph{Causal Analysis.} To understand the reason for the inconsistency in the causal effect, we analyze its relationship with problem complexity. In Table \ref{qa_causal}, we report the indirect effect of CoT with respect to the number of reasoning steps for GSM8K problems. We observe that with the increase in the number of reasoning steps, the indirect effect drops for both LLaMa-3 and ChatGPT. It indicates that the length of the reasoning steps has an inverse effect on the faithfulness of these models. Table \ref{tab:q_casual_analysis} shows a few examples of different models' unfaithful reasoning over the chain of thought.
\begin{table}[h!]
\small
\centering
{
\begin{tabular}{lll}
\toprule
\textbf{No. RS}	& \textbf{LLaMa-3-70B} & \textbf{ChatGPT} \\
\midrule
2	& \textbf{66.92} & 65.6 \\ 
3	& 50.31 & \textbf{53.9} \\ 
4	& 43.2 & \textbf{61.2} \\ 
5	& 41.8 & \textbf{55} \\ 
6	& \textbf{37.5} & 21 \\
7	& 0 & 25 \\ 
\midrule
overall	& \textbf{57.2} & 56.3 \\ 
\bottomrule
\end{tabular}}
\caption{\textbf{Indirect Effect.} No. RS = Number of Reasoning Steps. IE: $Avg[Acc(Y_{00}) - Acc(Y_{01})]$ }\label{qa_causal}
\end{table} 


\textbf{FRODO Analysis.} To further understand the findings in \S\ref{sec:result}, we manually analyze the relevance of the $100$ CoT generated by SFT and DPO. We observed that SFT generates 74\% and 54\% relevant CoT, whereas DPO generated CoT 77\% and 59\% relevant for QuaRel and OBQA tasks, respectively. Further, we observe two types of errors made by SFT and DPO: (i) invalid reasoning steps– reasoning steps leading to incorrect answers and (ii) unnecessary steps – reasoning steps not containing enough information to support a correct answer. For DPO, we observed that 40\% of the errors are invalid reasoning steps, and 56\% are unnecessary steps. Table \ref{tab:frodo_qa} shows some examples of CoT generated by SFT and DPO, SFT generated CoT are incomplete or contradictory. 


\section{Related Work}
\textbf{Measuring Faithfulness CoT.} 
\citet{jacovi-goldberg-2020-towards} argued that obtaining \textit{faithful} explanations that accurately reflect a model's reasoning process is important to understand the reasons behind its answer.\citep{atanasova-etal-2023-faithfulness} proposed a new benchmark to test the faithfulness of natural language explanations. \citet{Turpin2023LanguageMD} proposed identifying examples of unfaithful CoT in adversarial settings, showing that CoT reasoning is not always faithful. To determine faithfulness, they provided bias features in the few-shot setting or made edits to the input. \citep{Lanham2023MeasuringFI} argued that LLM ignores mistakes when introduced into the CoT, which reveals that the LLM is unfaithful. Finally, \citep{parcalabescu2023measuring} introduced CC-SHAP to measure input alignment with predictions for both post-hoc and CoT explanations. Unlike prior work, we employ causal mediation analysis to measure the model's faithful reasoning over the CoT, and to interpret its relationship with the answer.

\begin{table}[t!]
\centering
\footnotesize
\scalebox{1.0}{
\begin{tabular}{lcc}
\toprule 
  \textbf{Models}
 & \textbf{OBQA → SQA}
 & \textbf{QASC → SQA} \\
 
\midrule
\textbf{T5-3B + CoT}  & 67.6  & 53.2 \\ 
\textbf{\ourmodel{}}  &  69.4 &  56.2 \\ 
\bottomrule
\end{tabular}}
\caption{\textbf{Generalization Performance} (accuracy) of methods, trained on a source dataset and directly predicting on a target dataset (denoted as source → target).} \label{tab:generalization}

\end{table} 
\paragraph{Self-Rationalization and CoT Distillation.} Initial work on self-rationalization approaches focused on collected gold human rationales and training a model to learn to generate such rationales \citep{wiegreffe-etal-2021-measuring, paul-frank-2021-coins, NIPS2018_8163}. With the advent of LLMs, recently many works have distilled CoT from LLMs and endowed small LMs with step-by-step reasoning capabilities \citep{pmlr-v202-fu23d, LI2022ExplanationsFL, shridhar-etal-2023-distilling, li-etal-2023-symbolic}. Our work involves distilling CoT from LMs to a smaller one, similar to a certain line of work. We differ in using implicit feedback to enhance the correctness of the distilled CoT.

\paragraph{Feedback to Improve Reasoning.} Recently, several papers have proposed to improve or revise the LMs' generation using feedback \citep{fernandes2023bridging, Pan2023AutomaticallyCL}. Broadly, existing methods can be categorized into two kinds: external and intrinsic feedback. In the realm of external feedback, a standard procedure is to train critic models and use them to facilitate and improve the original generation model \citep{Peng2023CheckYF, Akyurek2023RL4FGN, Mehrabi2023FLIRTFL, Paul2023REFINERRF}. Among them, \citet{Paul2023REFINERRF} is related to our paper as it evaluates each reasoning step as feedback to produce more reasonable reasoning steps. In contrast to extrinsic feedback, which relies on external sources, there are works which show that internal knowledge of LLMs can be used to give feedback \citep{Kim2023LanguageMC, madaan2023selfrefine, Shinn2023ReflexionLA}. However, \citet{madaan2023selfrefine} argued that self-feedback does not improve performance on reasoning tasks. Hence, in this work, we create preference data (counterfactual and factual reasoning steps) to train a specialized model to learn to generate correct reasoning steps with implicit feedback. 

\paragraph{Casual Mediation Analysis in NLP.} Causal mediation analysis is an important tool that is used to effectively attribute the causal effect of mediators on an outcome variable \citep{Pearl2001DirectAI}. \citet{NEURIPS2020_92650b2e} proposed to use this method to implicate specific neurons and attention heads in mediating gender bias in various pre-trained LMs. Later, this method was used for analyzing different models' behaviour for different downstream tasks such as Subjective-Verb agreement \citep{Finlayson2021CausalAO}, Fake News Detection \citep{chen-etal-2023-causal}, arithmetic reasoning \citep{stolfo-etal-2023-mechanistic}, political polarization \citep{tierney-volfovsky-2021-sensitivity}. To the best of our knowledge, our study is the first attempt to use casual mediation analysis to analyze the faithfulness of LLMs in their reasoning capabilities. In this work, we followed \citet{Pearl2001DirectAI} to perform the mediation analysis. The mediation analysis allows us to measure the following: Direct effect: Contribution of X (input) to Y (output). Indirect effect: Contribution of R (reasoning chain) to Y (output). Hence, a high direct effect means the model's output is primarily influenced by the input, and a high indirect effect means the reasoning chain has more effect on the output.

\section{Conclusion}

In this work, we perform a causal mediation analysis to study the indirect effect of CoT on the final output of twelve LLMs. Our experiments show large variations across tasks and models in how strongly reasoning traces causally affect the model’s prediction. LLMs generally do not reliably use their intermediate reasoning steps when generating an answer. We introduce \ourmodel{} that tailors small-sized LMs to generate correct reasoning chains and faithfully reason over them to arrive at the correct answer. Experiments show that our method outperforms strong baselines on four reasoning tasks, including out-of-distribution settings.

\section*{Acknowledgment}

We would like to thank Angelika Romanou, Beatriz Borges, Sahithya Ravi, Gail Weiss, Maxime Peyrard, Syrielle Montariol, Anna Sotnikova, Negar Foroutan and Zeming Chen for their helpful feedback on a draft version of the paper. We acknowledge the support of the ICT-48 Network of AI Research Excellence Center “TAILOR” (EU Horizon 2020, GA No 952215). West's lab is partly supported by grants from the Swiss National Science Foundation (200021\_185043), Swiss Data Science Center (P22\_08), H2020 (952215), Microsoft Swiss Joint Research Center, and Google, and by generous gifts from Facebook, Google, and Microsoft. Antoine Bosselut gratefully acknowledges the support of the Swiss National Science Foundation (No. 215390), Innosuisse (PFFS-21-29), the EPFL Center for Imaging, Sony Group Corporation, and the Allen Institute for AI.


\newpage 
\section{Limitations} 
A limitation of our Causal Analysis metric is that it does not evaluate the model's real internal reasoning process. Without a complete understanding of the pertaining data and models' internal working process, it is difficult to know whether or not the chain of thought is faithful to the reasoning process. In this study, we provide insist and evidence that could explain how the model uses CoT. For future research, the causal mediation metric can be useful for measuring the extent to which new methods improve faithfulness. 
Compared to training a standard CoT distillation process, our method requires (i) additional counterfactual data generated by LLMs, which can be expensive, and (ii) training time increases as training Direct Preference Optimization is a two-step process. To manage the complexity of our already large-scale experiments involving (a) four different reasoning tasks, and (b) hyperparameter search grids, we ran experiments with 3 random seeds.  Additionally, \ourmodel{} is dependent on rationales generated by LLMs. Extra care should be taken when applying our model in production environments, especially when making critical decisions or exposing its generated contents directly to human end users.

\newpage

\bibliography{anthology, custom}

\clearpage

\appendix
\section{Appendix}

\subsection{FRODO - Inference Module (DPO)}\label{label:dpo}

The preference data of human or artificial annotators is modeled by a learnable implicit reward model $f_\theta$ under Bradley-Terry theories \citep{Bradley1952RANKAO}: 

\begin{equation}
     \pi_\theta(r_w > r_l |x) = \sigma(f_\theta(r_w, x) - f_\theta(r_l,x))
\end{equation}
where $\sigma$ is the sigmoid function. To learn $f_\theta$, DPO adopts a binary classification loss: 
\begin{equation}
        L_\text{DPO} = -\mathbb{E}_{\{x,r_w>r_l\}}\text{log}\sigma(f_\theta (r_w,x)  - f_\theta(r_l,x))
\end{equation}

The latent function $f_\theta$ is parameterized by the log-likelihood ratio between $\pi_\theta$ and $\pi_\text{sft}$:
\begin{equation}
    f_\theta(x,r) = \beta\text{log}\frac{\pi_\theta(r|x)}{\pi_\text{sft}(r|x)} 
\end{equation}

where $\beta$ a linear coefficient for scaling $f_\theta$. This parameterization is appealing as it aligns the training of an implicit reward model $f_\theta$ closely with training an LM policy $\pi_\theta$. 

\subsection{Additional Experimental Results} 

\begin{figure}
    \centering
    \includegraphics[width=0.7\linewidth]{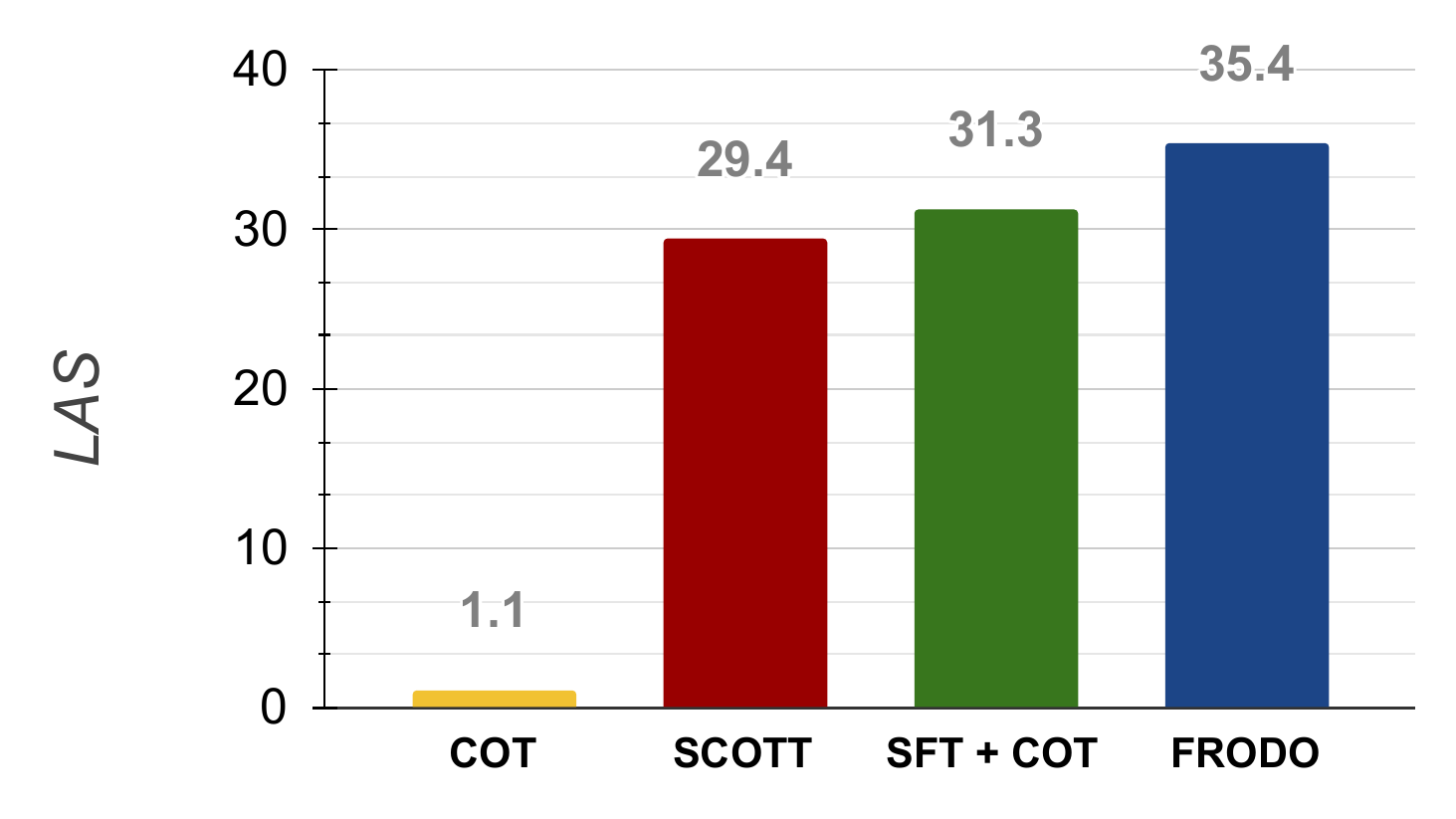}
    \caption{\textbf{Faithfulness} (LAS) of the compared methods on StrategyQA. The base Model is T5-3B.}
    \vspace*{-5mm}
    \label{fig:las}
\end{figure}

\paragraph{RQ1: How faithful FRODO is compared to SCoTT and CoT?} Finally, we compare the faithfulness of reasoning chains generated by \ourmodel{} with \textsc{SCoTT}, \textsc{CoT} and \textsc{SFT+COT} (see Fig.\ref{fig:las}). We observed that \ourmodel{} achieves a much higher LAS score than the other three baselines, suggesting that DPO training with implicit casual feedback helped the model. 

\paragraph{RQ2: How does FRODO work on GSM8k?} Table \ref{mathreasoning} reports the performance of FRODO on math reasoning problems. We observe that FRODO outperform SFT by +$3.75$ \% (average). 
\begin{table}[h]
\small
\centering
{
\begin{tabular}{ll}
\toprule
\textbf{Model} &	\textbf{GSM8K} \\
\midrule
LLama-2 7B + SFT + COT &	17.8 \\ 
LLama-2 7B + SFT + FRODO &	\textbf{21.1} \\ 
\midrule
Mistral + SFT +  COT &	40.4 \\ 
Mistral  + SFT + FRODO &	\textbf{44.6} \\ 
\bottomrule
\end{tabular}}
\caption{Performance of FRODO on GSM8K (accuracy)\label{mathreasoning}}
\end{table} 

\paragraph{RQ3: How does FRODO generalize on Causal Understanding Task?} Table \ref{causalunderstanding} reports the result of FRODO on the Causal Understanding task. Please note that the Causal Understanding dataset does not have training data. Hence, we trained FRODO on the StrategyQA dataset and evaluated it on Causal Understanding. Our results suggest that FRODO can generalize better than SFT. 

\begin{table}[h]
\small
\centering
{
\begin{tabular}{ll}
\toprule
\textbf{Model}	& \textbf{SQ → CU} \\
\midrule
SFT + COT	& 51.0 \\ 
SFT + FRODO	& \textbf{53.2} \\
\bottomrule
\end{tabular}}
\caption{\textbf{Generalization Performance.} (accuracy) of methods, trained on a source dataset and directly predicting on a target dataset (denoted as source → target), where SQ = StrategyQA and CU = Causal Understanding.}\label{causalunderstanding}
\end{table} 

\textbf{RQ4: How well did GPT-4 generate the Chain-of-Thought?} We manually evaluated the quality of the chain of thought generated by GPT-4. We found 94\% correct CoTs generated by GPT-4 for StrategyQA, whereas only 65\% correct for the Causal Understanding task. This also indicates why the performance of GPT-4 drops for the Causal Understanding task (see Table \ref{tab:closed_models_causal_resuts}). 
\begin{table}[ht]
\centering
{
\begin{tabular}{lll}
\toprule
\textbf{SQ}	& \textbf{GSM8K} & \textbf{CU} \\
94\% & 82\% & 65\% \\
\bottomrule
\end{tabular}}
\caption{Human Judgement of GPT-4 reasoning chain. SQ: StrategyQA, CU: Causal Understanding.}
\end{table} 

\paragraph{RQ5: What are the differences between FRODO and Selection-Inference \citep{creswell2023selectioninference} method?}

The key differences between FRODO and the Selection-Inference method are: 
\begin{enumerate}
    \item The selection-inference framework assumes that each question is accompanied by context information, which contains all the information necessary to solve the problem.
    \item  FRODO does not have that assumption; therefore, our method works on open-domain question-answer tasks. Hence, we compare our method with RAINER, CRYSTAL, and MARIO, which do not have such assumptions. 
    
\end{enumerate}

In the selection-inference framework, SFT with language modelling loss is used to train the inference module, while we used counterfactual loss, LM loss, and margin-ranking loss. 

\paragraph{RQ6: How is the performance of FRODO on the Entailmentbank dataset?}

Table \ref{entailmentbank} shows that FRODO clearly outperforms previous baselines on the Entailmentbank dataset.

\begin{table}[ht]
\centering
{
\begin{tabular}{lll}
\toprule
\textbf{Models}	& \textbf{Task 1} & \textbf{Task 2} \\
\midrule
Entailment Writer &	34.4 &	23.2\\
METGEN  &	37.0 &	28.0 \\
FRODO &	38.8 &	34.5\\
\bottomrule
\end{tabular}}
\caption{Performance of FRODO on Entailment Bank dataset.}\label{entailmentbank}
\end{table} 

\paragraph{RQ7: What is the causal effect of open-sourced models (<100B) in natural setting?} 
We experimented with open-sourced larger models LLaMa-3-70B (zero-shot setting) for natural indirect effect. However, we have observed that 70B models do not have the emergent abilities to generate coherent/meaningful reasoning steps in zero-shot settings. Similar to our observation, previous studies \citep{NEURIPS2022_8bb0d291} also discussed that only models >100B start showing such emergent abilities. Table \ref{llama-3-nie} reports the Natural Indirect effect of LLaMA-2 and LLaMA-3 (70B).

\begin{table}[ht]
\small
\centering
{
\begin{tabular}{lll}
\toprule
\textbf{Models}	& \textbf{Causal Effects} & \textbf{StrategyQA} \\ 
\midrule 
LLaMa-3 70B &	NIE & 21.1 \\
LLaMa-3 70B &	CIE	& 23.1 \\
LLaMa-2 70B &	NIE	& 12.1 \\ 
LLaMa-2 70B & 	CIE	& 24.1 \\ 
\bottomrule
\end{tabular}}
\caption{Performance of FRODO on GSM8K (accuracy)\label{llama-3-nie}}
\end{table}

\subsection{Dataset and Implementation Details}\label{sec:appendix_data_details}
All datasets have multi-choice questions ``\textit{yes}/\textit{no}'' for \textsc{StrategyQA}, ``\textit{a}/\textit{b}'' for \textsc{Quarel}, ``a/b/c/d'' for \textsc{OpenBookQA}, ``\textit{a}/\textit{b}/-/\textit{h}'' for \textsc{QASC}), and the task is to generate a rationale followed by the predicted answer. We use the original data splits (see Table.\ref{tab:dataset_details}). 
\begin{table}[t!]
\centering
\footnotesize
\begin{tabular}{lcc}
\toprule  
\textbf{Data Size} & \textbf{Test Data Size} \\
\midrule
\textbf{GSM8K}  & 300 \\
\textbf{Causal Understanding}  & $175$  \\
\textbf{StrategyQA}  & $500$ \\ 
\midrule
\end{tabular}
\caption{Data Statistics: Causal Mediation Analysis}\label{tab:training_details}
\end{table}

\begin{table}[t!]
\centering
\footnotesize
\begin{tabular}{lcc}
\toprule  
\textbf{Hyperparameter} & \textbf{Value} \\
\midrule
\textbf{Optimizer}  & Adam \\
\textbf{Adam epsilon}  & $1e-8$  \\
\textbf{Adam initial learning-rate}  & $3e-5$ \\ 
\textbf{Learning-rate scheduler} & linear with warmup \\
\textbf{Warmup steps} & $1000$ \\ 
\textbf{Gradient clipping} & $0.5$\\
\textbf{Train batch-size} & $4/8$\\
\textbf{Training Time} & $\sim 4$ hours on 1 GPU \\

\midrule
\end{tabular}
\caption{Training Details for small LMs}\label{tab:training_details}
\end{table}

\begin{table}[t!]
\centering
\footnotesize
\begin{tabular}{lcc}
\toprule  
\textbf{Hyperparameter} & \textbf{Value} \\
\midrule
\textbf{Optimizer}  & RMSprop \\
\textbf{Adam epsilon}  & $1e-8$  \\
\textbf{Train batch-size} & $4/8$\\
\textbf{beta} & 0.25 \\
\textbf{Training Time} & $\sim 8$ hours on 1 GPU \\
\midrule
\textbf{LoRA parameters} \\
\midrule
\textbf{task type} & CAUSALLM \\ 
\textbf{r} & 16 \\ 
\textbf{lora alpha} & 32 \\ 
\textbf{lora dropout} & 0.05 \\ 
\bottomrule
\end{tabular}
\caption{Training Details for Direct Preference Optimization}\label{tab:training_details}
\end{table}

\begin{table*}[h]
\centering
{
\scalebox{0.65}{
\begin{tabular}{@{}ll}
\toprule
{\bf Example 1}\\
\midrule
\textbf{Question:} Can members of the Green Party of England and Wales vote in the USA? \\
\textbf{Correct Answer:} \textbf{False} \\ 
\textbf{Gold Rationales:} Green Party of England Wales isn't registered in the USA. People who live in England can't vote in the USA.\\
\textbf{Counterfactual Rationales:} The Green Party of England and Wales was registered in the USA. People who live in England can be able to vote in the USA.\\
\textbf{GPT-4 Rationales:} Step 1: Understand that the Green Party of England and Wales is not registered in the USA, \\ meaning it doesn't have a direct influence or participation in the US elections.  Step 2: Consider the fact that people \\ who live in England, regardless of their party affiliation, are not eligible to vote in the USA. \\
\textbf{GPT-4 Answer ($Y_{00}$):} False \checkmark \\
\textbf{GPT-4 Answer ($Y_{01}$):} \ul{False} \xmark \\
\textbf{ChatGPT Rationales:} 1. The ability for members of a political party to vote in another country is typically determined by the registration \\ and recognition of that party in the respective country.  2. If the Green Party of England and Wales is registered and recognized in the USA, \\ it is possible that members of the party residing in England may have the opportunity to vote in the USA. \\
\textbf{Correct Answer:} False \\ 
\textbf{ChatGPT Answer ($Y_{00}$):} \uline{False} \checkmark \\ 
\textbf{ChatGPT Answer ($Y_{01}$):} \ul{False} \xmark \\
\textbf{LLaMa-3-70B Answer ($Y_{00}$):} False \checkmark \\
\textbf{LLaMa-3-70B Answer ($Y_{01}$):} True  \checkmark \\
\midrule
{\bf Example 2}\\
\midrule
\textbf{Question:} Could a wandering albatross fly from Bucharest to New York City without a rest?  \\
\textbf{Correct Answer:} \textbf{True} \\ 
\textbf{Gold Rationales:} Wandering albatross can travel at least 15,000 km (just under 10,000 miles) over the sea before returning to land. \\ It's 4766 miles or 7670 km from Bucharest to New York City.\\
\textbf{Counterfactual Rationales:}  If the distance from Bucharest to New York City was more than 15,000 km, \\ a wandering albatross would not be able to fly this distance without a rest.\\ 
\textbf{GPT-4 Rationales:} Step 1: Compare the distance a wandering albatross can travel before returning to land (15,000 km) with the \\ distance from Bucharest to New York City (7670 km).  Step 2: Since 7670 km is less than 15,000 km,  it can be concluded \\ that a wandering albatross could fly from Bucharest to New York City without a rest. \\
\textbf{GPT-4 Answer ($Y_{00}$):} True \checkmark \\ 
\textbf{GPT-4 Answer ($Y_{01}$):} {False} \checkmark \\
\textbf{ChatGPT Rationales:} 1. Determine the maximum distance the wandering albatross can fly without a rest: Given that the wandering albatross can travel \\ at least 15,000 km (just under 10,000 miles) over the sea before returning to land, we need to compare this distance to the distance between Bucharest and \\ New York City. 2. Calculate the distance between Bucharest and New York City: Given that it is 4766 miles or 7670 km from Bucharest to New York City. \\
\textbf{ChatGPT Answer ($Y_{00}$):} True \checkmark \\ 
\textbf{ChatGPT Answer ($Y_{01}$):} {False} \checkmark \\
\textbf{LLaMa-3-70B Answer ($Y_{00}$):} True \checkmark \\
\textbf{LLaMa-3-70B Answer ($Y_{01}$):} \ul{True} \xmark \\
\bottomrule
\end{tabular}}}
\caption{Qualitative Examples of Causal Mediation Analysis. }\label{tab:q_casual_analysis}
\end{table*}

\begin{table*}[h]
\small
\centering
{
\begin{tabular}{@{}lll@{~~~}l@{}}
\toprule
{\bf Dataset/Tools} & {\bf Citation}& {\bf Link} & \bf{License}\\
\midrule
GSM8k & \citet{cobbe2021gsm8k} & \url{https://github.com/openai/grade-school-math} & {MIT License}\\
HuggingFace  & \citet{wolf-etal-2020-transformers} & \url{https://github.com/huggingface/transformers} & {Apache License}\\
OBQA & \cite{mihaylov-etal-2018-suit} & \url{https://huggingface.co/datasets/openbookqa} &  {Apache License}\\
StrategyQA & \cite{geva-etal-2021-aristotle}  & [1] & {MIT License}\\
Quarel & \cite{Tafjord2018QuaRelAD} & \url{https://github.com/allenai/unifiedqa} & {MIT License}\\
QASC & \citep{khot2020qasc} & \url{https://github.com/allenai/unifiedqa} & {MIT License}\\

\bottomrule
\end{tabular}}
\caption{More details about datasets and Tools [1] \url{https://github.com/eladsegal/strategyqa/tree/main/data/strategyqa}}\label{tab:dataset_details}
\end{table*}

\begin{table*}[h]
\centering
{
\scalebox{0.60}{
\begin{tabular}{@{}llll@{~~~}l@{}}
\toprule
{\bf Dataset} & {\bf Question }& {\bf Option }& {\bf Correct Answer} & \bf{GPT-3 Generated CoT}\\
\midrule
\textbf{StrategyQA} & Can I build a house on an asteroid?  & Yes or No & No & Building a house on an asteroid is  \\ &&&&impossible as of now due  \\ &&&&  to the lack of technologies \\ &&&& and resources needed. \\ &&&& It would be extremely difficult \\ &&&& to build a house that could \\ &&&& withstand the extreme \\ &&&& temperatures, radiation, \\ &&&&  and extreme gravitational pull. \\
\midrule
\textbf{OBQA} & The circulatory system brings & (a) The brain (b) The feet & The chest &  The circulatory system brings oxygen \\ &oxygen to the body from where? &(c) The stomach area && to the body from the lungs \\ &&(d) The chest & & which is located in the chest area. \\
\midrule
\textbf{Quarel} & The boys were racing their & (A) weighed more & weighed less & When something is lighter, \\ & cars in the soapbox derby and found that the cars & (B) weighed less & & it is easier to move faster. \\ & that $----$  moved faster.  &   && Thus, the cars that weighed less moved faster.\\
\midrule
\textbf{QASC} & What type of water formation is formed by clouds?  & (A) pearls (B) streams (C) shells & Beads & Rain is formed when water droplets \\ && (D) diamonds (E) rain (F) beads && in the clouds come together to form larger \\ &&(G) cooled (H) liquid && droplets that are too heavy to remain\\ &&&& suspended in the cloud, and fall to \\ &&&& the ground as precipitation.\\
\bottomrule
\end{tabular}}}
\caption{Examples from each reasoning task.}\label{tab:dataset_examples}
\end{table*}

\begin{table*}
\centering
\centering
\begin{tabular}{l}
\toprule
\textbf{PROMPT: Counterfactual Reasoning Chain} \\
\midrule
\texttt{System Prompt: You are a helpful assistant for commonsense reasoning.} \\
\texttt{We will provide you with a commonsense question, along with a correct} \\
\texttt{answer and your task is to generate a counterfactual intermediate.} \\
\texttt{steps.  Here are two examples: } \\ \\
\texttt{“Question : ” <Problem Statements> Let’s think step by step }\\
\texttt{Answer: <answer> } \\ \\
\texttt{“Question: ” <Problem Statements> Let’s think step by step }\\
\texttt{Answer: <answer> } \\ \\
\texttt{“Question: ” <Problem Statements> Let’s think step by step  }\\
\bottomrule
\end{tabular}
\caption{Prompts used for generating counterfactual intermediate reasoning chains.}
\label{table:counterfactual_prompt}
\end{table*}

\begin{table*}
\centering
\centering
\begin{tabular}{l}
\toprule
\textbf{PROMPT: Counterfactual Questions} \\
\midrule
\texttt{System Prompt: You are a helpful assistant in generating counterfactual questions.} \\
\texttt{We will provide you with a commonsense question, along with a correct} \\
\texttt{answer and your task is to generate a counterfactual question.} \\
\texttt{Here are two such examples: } \\ \\
\texttt{“Question : ” <Original Reasoning Question>  " Answer: "  <answer>}\\
\texttt{"Counterfactual Question:": <counter question> } \\ \\
\texttt{“Question : ” <Original Reasoning Question> " Answer: "  <answer>}\\
\texttt{"Counterfactual Question:": <counter question> } \\ \\
\texttt{“Question : ” <Original Reasoning Question>  " Answer: "  <answer>}\\
\texttt{"Counterfactual Question:": } \\
\bottomrule
\end{tabular}
\caption{Prompts used for generating counterfactual reasoning questions.}
\label{table:counterfactual_question}
\end{table*}

\subsection{Baselines}\label{sec:appendix_baselines}
We evaluate a series of language models that are diverse in terms of scale, training, and data:
\begin{itemize}
[leftmargin=*,topsep=1pt,parsep=1pt]
    \item \textbf{LLaMA} \cite{Touvron2023Llama2O}, an open-source decoder-only model with various sizes (7B) model is pretrained using only a language modeling loss.
    \item \textbf{GPT-3.5} \cite{NEURIPS2020_gpt3} and \textbf{GPT-4} \cite{openai2023gpt4}: two closed-source decoder-only models that were trained with instruction-tuning. For GPT-3.5, we use the \texttt{text-davinci-003} model with 175B parameters. 
    \item \textbf{Stable-Vicuna}: 
      open-source decoder-only model based on LLaMA. Stable-Vicuna is fine-tuned with RLHF.
    \item \textbf{Flan-T5-XXL} (\citealp{chung2022scaling}, 11B parameters) and \textbf{Flan-Alpaca} (\citealp{chia2023instructeval,peng2023instruction}; 3B), two open-source encoder-decoder models based on T5 \cite{raffel2020exploring} and trained on instruction-following datasets.
    \item \textbf{Mistral} \citep{jiang2023mistral} a 7–billion-parameter language model and \textbf{Mixtral} \citep{jiang2024mixtral} a Sparse Mixture of Experts (SMoE) language model. Mixtral has the same architecture as Mistral 7B, with the difference
that each layer is composed of 8 feedforward blocks (i.e. experts).
\end{itemize}

\subsection{Details about Preference Data}  
In our experiments, we consider two kinds of reasoning chains incorrect: counterfactual chains (alternative chains that can lead to different outcomes) and irrelevant chains (irrelevant facts about the correct answer). Additionally, we train DPO with a setting where for each instance, one correct and one incorrect chain are paired and given to the model. The model learns to prefer the correct chain. In Table \ref{table:preference_data}, we report the size of the preference data used to train the DPO models.

\begin{table*}[h!]
\centering
\footnotesize
\begin{tabular}{ll}
\toprule  
\textbf{Examples} & \textbf{Generations} \\
\midrule
\textbf{Question}  & Is the Illuminati card game still popular? \\
\textbf{Gold Reasoning Chain}  &  The original version of the game was released in 1982. \\
& A collectible card game version was released in 1995 but only had one set. \\
& The most recent edition of the base game was published in 2007. \\
\midrule
\textbf{SFT + CoT}  &  The Illuminati card game was released in the 1980s. (\textcolor{red}{\textbf{Incomplete}})  \\

\textbf{DPO} & The Illuminati card game was released in the 1980s. \\ &  The Illuminati card game was discontinued in the 1990s. \\
\midrule
\textbf{Correct Answer} & False \\
\textbf{SFT} & \ul{True} \xmark \\
\textbf{FRODO} & False \checkmark \\
\midrule
\textbf{Question}  & Tank the kitten learned from trial and error that carpet is rougher \\ & then skin. When he scratches his claws over carpet it generates \\ &  ---- then when he scratches his claws over skin (A) more heat (B) less heat \\
\textbf{GPT-3 Reasoning Chain}  &  When a cat scratches its claws over a rough surface, \\  & it generates more heat than when it \\ & scratches its claws over a smooth surface.\\
\midrule
\textbf{SFT + CoT}  &  When you scratch a surface, it generates heat. \\ & When you scratch a surface, it generates less heat. (\textcolor{red}{\textbf{Contradiction}}) \\

\textbf{DPO} & When a cat scratches a surface, it generates heat.  \\ 
\midrule
\textbf{Correct Answer} & A \\
\textbf{SFT} &  A \checkmark\\
\textbf{FRODO} &  A \checkmark \\
\midrule 

\end{tabular}
\caption{Qualitative Examples of model generated rationales and prediction. }\label{tab:frodo_qa}
\end{table*}

\begin{table*}
\centering
\centering
\begin{tabular}{lllll}
\toprule
Data type &	StrategyQA &	QuaRel	 & OBQA &	QASC \\
\midrule
Correct Reasoning Chain ($R_w$)	&5492	&8203	&20138	&19935 \\
\midrule
Counterfactual Reasoning Chain ($R_l$)	&5492	&8203	&20138	&19935 \\ 
Irrelvant Reasoning Chain ($R_l$)	& 5492	&8203	&20138	& 19935 \\
\bottomrule
\end{tabular}
\caption{Preference Data Statistics.}
\label{table:preference_data}
\end{table*}


\begin{table*}[h!]
\centering
\footnotesize
\begin{tabular}{lcc}
\toprule  
\textbf{Tasks} & \textbf{Interventions} \\
\midrule
\textbf{StrategyQA}  & Prompt GPT-4 to generate alternative questions such that \\ 
& the answer changes from original to counterfactual. \\
\midrule 
\textbf{GSM8K}  & We automatically replace the operands with alternative operands.  \\
\midrule 
\textbf{Causal Understanding} & Prompt GPT-4 to generate alternative questions such that \\
& the answer changes from original to counterfactual. \\
\midrule
\end{tabular}
\caption{Causal Interventions}\label{tab:causal interventions}
\end{table*}




\end{document}